\documentclass{article}

\usepackage{arxiv}
\usepackage[utf8]{inputenc}
\usepackage[T1]{fontenc}
\usepackage{hyperref}
\usepackage{url}
\usepackage{booktabs}
\usepackage{amsfonts}
\usepackage{microtype}
\usepackage{amsmath}
\usepackage{bbm}
\usepackage{natbib}
\usepackage{doi}
\usepackage[tableposition=below]{caption}
\usepackage{spverbatim}
\usepackage{tikz}
\usepackage{multirow}
\usepackage{graphicx}

\usetikzlibrary{shapes.geometric, arrows.meta, positioning}

\title{UtilityMax Prompting: A Formal Framework for Multi-Objective Large Language Model Tasks}

\date{}

\newif\ifuniqueAffiliation
\uniqueAffiliationtrue
\author{
Ofir Marom \\
Independent Researcher \\
\texttt{ofiremarom@gmail.com}
}

\renewcommand{\headeright}{Technical Report}
\renewcommand{\undertitle}{Technical Report}
\renewcommand{\shorttitle}{UtilityMax Prompting}

\begin{document}
\maketitle

\begin{abstract}
The success of a Large Language Model (LLM) task depends heavily on its prompt. Most use-cases specify prompts using natural language, which is inherently ambiguous when multiple objectives must be simultaneously satisfied. In this paper we introduce \textbf{UtilityMax Prompting}, a framework that specifies tasks using formal mathematical language. We reconstruct the task as an influence diagram in which the LLM's answer is the sole decision variable. A utility function is defined over the chance nodes of the diagram, and the LLM is instructed to find the answer that maximises expected utility. This constrains the LLM to reason explicitly about each component of the objective, directing its output toward a precise optimization target rather than a subjective natural language interpretation. We validate our approach on the MovieLens 1M dataset across three frontier models (Claude Sonnet 4.6, GPT-5.4, and Gemini 2.5 Pro), demonstrating consistent improvements in precision and Normalized Discounted Cumulative Gain (NDCG) over natural language baselines in a multi-objective movie recommendation task.
\end{abstract}

\section{Introduction}\label{sec:introduction}

Prompt engineering is crucial in directing Large Language Models (LLMs) to solve specific tasks. Various prompt engineering techniques have been proposed in recent years \cite{sahoo2024}. In particular, zero-shot prompting is attractive due to its simplicity since it does not require external exemplars to construct the prompt \cite{li2023, radford2019}.

One line of research focuses on restructuring how the model reasons within a single inference pass. With zero-shot Chain-of-Thought (CoT) prompting, the LLM is instructed to reason ``step-by-step'' about the solution before producing an answer \cite{wei2022, cheng2025}. In the few-shot CoT setting, exemplars of similar questions and answers are provided so that the LLM can observe the reasoning patterns it should follow \cite{wei2022}. In Program of Thoughts (PoT) prompting, the LLM translates a natural language prompt into a programming language such as Python, which is then executed to produce the final answer \cite{chen2023}. In Chain-of-Symbol (CoS) prompting, a task is converted into a symbolic representation that assists the LLM in solving spatial reasoning and planning tasks \cite{hu2023}.

A second line of research treats the prompt itself as the optimization target. For example, Optimization by Prompting (OPRO) uses an LLM as an iterative optimizer, generating candidate prompts, evaluating them against a labeled dataset, and refining them across multiple passes \cite{yang2024}. While effective, this approach requires a scoring function for the task so that proposed answers can be evaluated as better or worse. Such evaluation signals are not always available, or are expensive to obtain in many real-world deployment settings. 

These methods have demonstrated significant improvements in LLM performance across a range of benchmarks. However, they share a common assumption: the task objective is specified in natural language. While this is not a limitation in domains with a single objective, such as solving mathematical problems, it becomes a challenge when multiple dependent objectives must be optimized simultaneously.

For example, consider a trading agent that wants to maximise profit at a given level of risk tolerance. These objectives naturally compete with each other as the most profitable strategy may require taking a level of risk that could result in significant losses. Balancing both objectives therefore becomes key, but a natural language prompt that says: ``maximise profit subject to a medium level of risk'' is inherently ambiguous. The LLM has to interpret what ``medium'' means here. While a more carefully crafted natural language prompt could partially resolve this kind of ambiguity, a task like this is best specified in formal mathematical language that eliminates ambiguity entirely.

In this paper we introduce UtilityMax Prompting, a zero-shot framework that takes a complementary approach: rather than restructuring the reasoning process or iterating over candidate prompts, we replace the natural language objective itself with a formal mathematical specification. The LLM is instructed to maximise the expected utility given its answer, constraining it to reason about each component of the objective individually and directing its output toward a well-defined optimization target. This approach requires neither exemplars nor a scoring function. 

The rest of this paper is organized as follows: in Section \ref{sec:framework} we cover the general methodology behind the UtilityMax framework; in Section \ref{sec:prompting} we provide guidelines as well as a prompting template under UtilityMax for practitioners; in Section \ref{sec:experiments} we validate UtilityMax on a multi-objective movie recommendation task using the MovieLens 1M dataset \cite{harper2015}, demonstrating consistent improvements in precision and Normalized Discounted Cumulative Gain (NDCG) over natural language baselines across three frontier models (Claude Sonnet 4.6, GPT-5.4, and Gemini 2.5 Pro); and we conclude with final remarks and directions for future research in Section \ref{sec:final-remarks}.

\section{Framework}\label{sec:framework}

We formalise UtilityMax as an influence diagram in which the LLM's answer is the sole decision variable.

Let $K$ represent the LLM's knowledge. This includes all internal knowledge stored through its parameters as well as external knowledge such as context through chat history or online research. Let $A \mid K$ be a decision node that represents the space of all possible LLM answers given $K$. An answer $a \in A$ is decided by the LLM. Let $\mathbf{X} = \{X_1, X_2, \ldots, X_n\}$ be a set of chance nodes with $n \geq 1$. The dependency structure over $\{A\} \cup \mathbf{X}$ forms a DAG with $A$ as the singular root node. Let $U: \text{supp}(\mathbf{X}) \to \mathbb{R}$ be a utility function defined over the chance nodes.

UtilityMax is then specified by the tuple $(\text{DAG}, U)$. The DAG and the utility are coupled: the choice of $U$ determines what the LLM must compute, and the DAG determines whether that computation is tractable. Specifically, UtilityMax requires that the expected utility decompose as:

\begin{equation}\label{eq:decomposition}
\mathbb{E}[U \mid A] = \sum_{k=1}^N \prod_{i=1}^n \mathbb{E}[f_{i,k}(X_i) \mid A].
\end{equation}

This sum-of-products form is what the framework computes: each $\mathbb{E}[f_{i,k}(X_i) \mid A]$ is a single-variable conditional expectation that the LLM estimates independently, and these components are composed by the surrounding sum and product structure.

The task of the LLM is to find $a^* \in A$ such that $\mathbb{E}[U \mid A=a^*]$ is maximised.

The boundary of the framework is therefore the family of $(\text{DAG}, U)$ configurations whose expected utility admits this decomposition. Sections~\ref{ssec:conditional-independence} and~\ref{ssec:binary-gating} present two such configurations: conditional independence and binary gating. Each section specifies the DAG and the corresponding tractable forms of $U$ together, since the two are coupled.

\subsection{Conditional Independence}\label{ssec:conditional-independence}

In the simplest configuration, we assume the chance nodes are mutually conditionally independent given the LLM's answer.

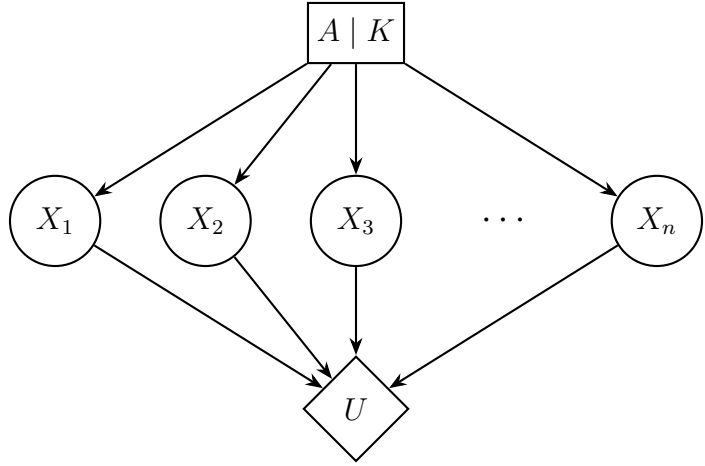
\begin{figure}[htbp]
\centering
\begin{tikzpicture}[
decision/.style={
rectangle, 
draw=black, 
thick,
minimum width=1.2cm, 
minimum height=0.8cm,
font=\large
},
chance/.style={
circle, 
draw=black, 
thick,
minimum size=1.2cm,
font=\large
},
utility/.style={
diamond, 
draw=black, 
thick,
minimum width=1.4cm, 
minimum height=1.4cm,
font=\large,
aspect=1.5
},
arrow/.style={
-Stealth, 
thick
}
]

\node[decision] (A) at (0, 0) {$A \mid K$};

\node[chance] (X1) at (-4, -2.5) {$X_1$};
\node[chance] (X2) at (-2, -2.5) {$X_2$};
\node[chance] (X3) at (0, -2.5) {$X_3$};

\node at (2, -2.5) {\Large $\cdots$};

\node[chance] (Xn) at (4, -2.5) {$X_n$};

\node[utility] (U) at (0, -5) {$U$};

\draw[arrow] (A) -- (X1);
\draw[arrow] (A) -- (X2);
\draw[arrow] (A) -- (X3);
\draw[arrow] (A) -- (Xn);

\draw[arrow] (X1) -- (U);
\draw[arrow] (X2) -- (U);
\draw[arrow] (X3) -- (U);
\draw[arrow] (Xn) -- (U);

\end{tikzpicture}

\caption{Influence diagram for UtilityMax Prompting under conditional independence, showing decision node $A$, chance nodes $\{X_1, X_2,\ldots,X_n\}$, and utility node $U$.}
\label{fig:influence-diagram}
\end{figure}

\textbf{DAG.} Each $X_i$ is conditionally independent given $A$. This setup is shown visually in Figure~\ref{fig:influence-diagram}.

\textbf{Utility.} Under conditional independence, any utility function of the sum-of-products form:

\begin{equation}
U(\mathbf{X}) = \sum_{k=1}^N \prod_{i=1}^n f_{i,k}(X_i)
\end{equation}

for $N \geq 1$ admits the decomposition. Since each $X_i$ is conditionally independent given $A$, the expectation of each product term factorises and the expected utility reduces to:

\begin{equation}
\mathbb{E}[U \mid A] = \sum_{k=1}^N \prod_{i=1}^n \mathbb{E}[f_{i,k}(X_i) \mid A].
\end{equation}

Each component $\mathbb{E}[f_{i,k}(X_i) \mid A]$ is estimated independently by the LLM, regardless of the choice of $f_{i,k}$. Two common patterns include:

\paragraph{Multiplicative utility} ($N=1$). A single product across all components, where every objective must hold simultaneously:

\begin{equation}
\mathbb{E}[U \mid A] = \prod_{i=1}^n \mathbb{E}[f_i(X_i) \mid A].
\end{equation}

\paragraph{Additive utility} ($N=n$). A sum across single-component terms, where each objective contributes independently to the total:

\begin{equation}
\mathbb{E}[U \mid A] = \sum_{k=1}^n \mathbb{E}[f_k(X_k) \mid A].
\end{equation}

\subsection{Binary Gating}\label{ssec:binary-gating}

In the second configuration, we relax the conditional independence assumption and instead introduce a prerequisite dependency structure among the chance nodes.

\textbf{DAG.} If all internal (non-leaf) chance nodes are binary random variables, we can introduce dependencies between chance nodes while preserving the decomposition. Let $pa(X_i)$ denote the parents of $X_i$ in the DAG, and suppose that if any $X_j = 0$ for $X_j \in pa(X_i)$ then $P(X_i = 1 \mid pa(X_i), A) = 0$. That is, each internal node is deterministically gated by its parents so that a child node can only be active if all its parents are active. Leaf nodes are not required to be binary and may take continuous or categorical values.

Denote by $pa(X_i) = 1$ the event that $X_j = 1$ for all $X_j \in pa(X_i)$. Let $\mathbf{L} \subset \mathbf{X}$ denote the set of leaf nodes and $\mathbf{I} = \mathbf{X} \setminus \mathbf{L}$ denote the set of internal nodes. We additionally assume that the leaf nodes are conditionally independent of each other given their parents and $A$:

\begin{equation}
X_j \perp X_k \mid pa(X_j) = 1,\ pa(X_k) = 1,\ A \quad \text{for all } X_j, X_k \in \mathbf{L},\ j \neq k.
\end{equation}

This is a strictly weaker condition than the global conditional independence of Section~\ref{ssec:conditional-independence}, which requires every node to be independent given $A$ alone. A sufficient structural condition under which it holds automatically is that the DAG is tree-structured, since leaf nodes on distinct branches then share no common parent other than internal nodes that are fixed to one.

\textbf{Utility.} For internal nodes $X_i \in \mathbf{I}$ we set $f_i(X_i) = X_i$. This is without loss of generality since the gating semantics require $f_i(0) = 0$, and any positive scaling constant cancels in the $\operatorname{argmax}$. Under these assumptions, any utility of the form:

\begin{equation}
U(\mathbf{X}) = \sum_{k=1}^N \left( \prod_{i:\, X_i \in \mathbf{I}_k} X_i \cdot \prod_{j:\, X_j \in \mathbf{L}_k} f_j(X_j) \right),
\end{equation}

where $\mathbf{I}_k \subseteq \mathbf{I}$ and $\mathbf{L}_k \subseteq \mathbf{L}$ are arbitrary subsets, admits the decomposition. The corresponding expected utility is:

\begin{equation}
\mathbb{E}[U \mid A] = \sum_{k=1}^N \left( \prod_{i:\, X_i \in \mathbf{I}_k} P(X_i = 1 \mid pa(X_i) = 1, A) \times \prod_{j:\, X_j \in \mathbf{L}_k} \mathbb{E}[f_j(X_j) \mid pa(X_j) = 1, A] \right).
\end{equation}

Each component $P(X_i = 1 \mid pa(X_i) = 1, A)$ for $X_i \in \mathbf{I}$ and $\mathbb{E}[f_j(X_j) \mid pa(X_j) = 1, A]$ for $X_j \in \mathbf{L}$ is estimated independently by the LLM. As in Section~\ref{ssec:conditional-independence}, two common patterns include:

\paragraph{Multiplicative utility} ($N=1$, $\mathbf{I}_1 = \mathbf{I}$, $\mathbf{L}_1 = \mathbf{L}$).  A single product across all internal and leaf nodes:

\begin{equation}
\mathbb{E}[U \mid A]
= \prod_{i:\, X_i \in \mathbf{I}} P(X_i = 1 \mid pa(X_i) = 1, A)
\times
\prod_{j:\, X_j \in \mathbf{L}} \mathbb{E}[f_j(X_j) \mid pa(X_j) = 1, A].
\end{equation}

\paragraph{Path-sum utility} ($N = |\mathbf{L}|$, one term per leaf). For each leaf $X_j \in \mathbf{L}$, let $\mathbf{I}_j \subseteq \mathbf{I}$ denote the set of internal node ancestors of $X_j$. Each term in the sum corresponds to one leaf, weighted by the probability that all its prerequisites are met:

\begin{equation}
\mathbb{E}[U \mid A] = \sum_{j:\, X_j \in \mathbf{L}} \left( \prod_{i:\, X_i \in \mathbf{I}_j} P(X_i = 1 \mid pa(X_i) = 1, A) \times \mathbb{E}[f_j(X_j) \mid pa(X_j) = 1, A] \right).
\end{equation}

\section{Prompting}\label{sec:prompting}

\subsection{Variable Guidelines}\label{ssec:variable-guidelines}

The effectiveness of UtilityMax depends not only on the structure of the influence diagram but also on the precision of the variable definitions. There is an inherent trade-off: a precisely defined variable constrains the LLM to a narrow target, reducing ambiguity in its estimates but potentially missing relevant signal that falls outside the strict definition. A loosely defined variable gives the LLM more latitude to draw on its broader knowledge, at the cost of reintroducing the same ambiguity that UtilityMax is designed to eliminate.

The mechanism behind this trade-off is straightforward. The LLM estimates each $\mathbb{E}[f_i(X_i) \mid A=a]$ using its internal knowledge $K$, which is shaped by the events and outcomes described in its training data. Events that appear consistently in training data --- behavioural outcomes, measurable metrics, factual properties --- have a stable representation that the LLM can draw on. Events expressed through subjective or interpretive language --- ``happiness,'' ``quality,'' ``fit'' --- appear in training data with much more variability in how they are described, and the LLM has no canonical referent to estimate against. A variable definition is therefore most precise when it can be expressed in terms of the kinds of events the LLM's training data describes consistently.

The practical implication is that variables should be defined to describe an observable event whose outcome could in principle be measured. The test the designer can apply is: \textit{would this exact phrasing appear in a news article, product review, technical report, or scientific paper?} If yes, the LLM likely has a calibrated sense of the event. If not, the estimate will rely on natural-language interpretation and reintroduce the ambiguity the framework is designed to eliminate.

When the underlying concept is itself subjective or internal, the technique is to replace it with an observable correlate --- a downstream event that the subjective concept should predict, and which the LLM can reason about concretely. A few examples:

\begin{itemize}
\item \textbf{Subjective state to behavioural indicator.} ``The user is happy with the product'' is internal and difficult to estimate; ``the user opens the app at least three times in the next week'' is observable and behaviourally well-defined.
\item \textbf{Judgement to measurable outcome.} ``The candidate is a good fit for the role'' is interpretive; ``the candidate completes the probation period and remains employed after six months'' specifies a concrete future event.
\item \textbf{Quality to threshold of significance.} ``The recommendation is good'' is vague; ``the user rates the recommendation $4$ or higher on a $5$-point scale'' specifies the success criterion in terms the LLM can reason about directly.
\item \textbf{Outcome to operational definition.} ``Collaboration occurs'' is loose; ``at least one person contacts me within two weeks about running a joint experiment'' specifies the observable event whose probability the LLM is being asked to estimate.
\end{itemize}

In each case the move is the same: replace the subjective or interpretive concept with a concrete downstream observable that the original concept should predict. The LLM is then estimating the probability of an event it can reason about consistently, rather than the probability of an interpretation it has to construct.

The right level of precision depends on the task. When the outcome can be well-specified, a more precise definition is generally preferred. When the concept is inherently subjective and no useful observable correlate exists, some looseness may be acceptable --- but the designer should treat this as a signal that the variable is doing less work than it appears to and may not meaningfully discriminate between candidate answers.

\subsection{Utility Guidelines}\label{ssec:utility-guidelines}

Once the variables of the influence diagram have been defined, the designer must choose how they combine into a utility function. The form of $U$ should reflect how the objectives genuinely interact in the task at hand. UtilityMax supports this through a small set of compositional operations on the chance nodes that carry direct semantic meaning:

\begin{itemize}
\item Product is logical AND. A term $f_i(X_i) \cdot f_j(X_j)$ models the requirement that both objectives contribute simultaneously. If either is weak, the term collapses.
\item Sum is logical OR. A term $f_i(X_i) + f_j(X_j)$ models substitutable objectives. Each contributes additively to the total, and strength on one can compensate for weakness on another.
\item For binary $X_i$, choosing $f_i(X_i) = 1 - X_i$ is logical NOT. This allows the designer to express that an objective should fail rather than hold.
\item Scaling by a constant $c$, with $f_i(X_i) = c X_i$, is weighting. A positive weight expresses how much an objective matters relative to the others; a negative weight expresses a penalty when the objective holds. Scaling under multiplicative composition does not change the $\arg\max$ and so weighting is most meaningful under sum-containing compositions.
\end{itemize}

A continuous variable $X_i$ can be converted to a binary indicator by defining $f_i(X_i) = \mathbbm{1}[X_i > \tau]$ for some threshold $\tau$. The LLM estimates $P(X_i > \tau \mid \cdot)$ directly, and the indicator can then participate in AND, OR, or NOT compositions like any other binary component.

Under either configuration of Section~\ref{sec:framework}, any utility built from these operations is tractable and carries an explicit semantic reading that the designer can verify against the task. The four practical patterns introduced in Section~\ref{sec:framework} arise as instances of this grammar:

\begin{itemize}
\item \textbf{Multiplicative utility under conditional independence} is an all-AND composition: every objective must hold simultaneously, and a weak value on any single objective suppresses the entire utility.
\item \textbf{Additive utility under conditional independence} is an all-OR composition: objectives are substitutable, each contributes independently to the total, and trade-offs between them are acceptable.
\item \textbf{Multiplicative utility under binary gating} is a single AND-chain: every gating condition and every leaf contribution must hold simultaneously.
\item \textbf{Path-sum utility under binary gating} is OR across leaves with AND within each path: each leaf contributes independently to the total, but each leaf's contribution requires all of its prerequisite gates to hold.
\end{itemize}

\paragraph{Converting a natural-language objective.} The grammar provides the language for expressing utility structures, but a natural-language objective rarely arrives in a form that maps directly to a specific composition. The designer must first decompose the objective into its constituent clauses and classify each clause according to its role in the utility.

A natural-language objective is decomposed into atomic clauses, where each clause expresses a single condition that the answer should satisfy. Compound clauses connected by ``and'' are split; clauses connected by ``or'' or expressing alternatives are kept together as disjunctive groups; clauses expressing the absence of a condition are flagged as candidates for NOT.

Each clause is then classified as either a \textit{gate} or a \textit{quality contributor}:

\begin{itemize}
\item A clause is a \textbf{gate} if its failure should disqualify the candidate. If the condition fails entirely, the overall utility should collapse to zero, regardless of how well the remaining conditions are met. Hard prerequisites, structural requirements, and binary feasibility conditions are typically gates.
\item A clause is a \textbf{quality contributor} if its failure should reduce but not eliminate the candidate's score. If the condition fails entirely, the candidate should still receive credit for the conditions it does satisfy. Aspirational properties, soft preferences, and quality-improving features are typically quality contributors.
\end{itemize}

The classification can be settled by asking, for each clause: \textit{if a candidate fails entirely on this condition but holds strongly on every other, is the candidate still in contention?} If no, the clause is a gate. If yes, it is a quality contributor.

Once each clause is classified, the utility is composed as follows. Gates combine multiplicatively, so that any failed gate suppresses the entire utility. Quality contributors combine additively within a single term that itself sits multiplicatively alongside the gates:

\begin{equation}
U(\mathbf{X}) = \left( \prod_{i \in \mathbf{G}} f_i(X_i) \right) \cdot \left( \sum_{j \in \mathbf{Q}} f_j(X_j) \right),
\end{equation}

where $\mathbf{G}$ is the set of gate clauses and $\mathbf{Q}$ is the set of quality contributors. This form ensures that gate failure zeros the utility while allowing quality contributors to substitute for one another. The all-multiplicative utility is recovered when $\mathbf{Q} = \emptyset$, and the all-additive utility is recovered when $\mathbf{G} = \emptyset$ (with the empty product taken as $1$). A natural-language objective with predominantly gate-shaped clauses will produce a utility that is mostly multiplicative; an objective with predominantly quality-shaped clauses will produce one that is mostly additive. The classification procedure ensures that this choice reflects the actual structure of the objective rather than defaulting to one form.
\subsection{Template}\label{ssec:template}

Given the framework outlined in Section \ref{sec:framework}, we can construct a prompt to maximise the given utility. A template for such a prompt using multiplicative utility is given below for two random variables $X_1$ and $X_2$.

\begin{spverbatim}
I want you to solve the following task: [TASK DESCRIPTION].

Formally, let K represent your knowledge. This includes all your internal knowledge stored through your parameters as well as any external knowledge provided in this prompt or chat history.

Let P(A | K) represent your probability distribution over answers given K. Let a be an answer in A.

Let X1 | A=a be a random variable representing [DESCRIPTION OF X1] given answer a.

Let X2 | A=a be a random variable representing [DESCRIPTION OF X2] given answer a.

Your task is to use your domain expertise to find the optimal answer a* that maximises O(a) = E[X1 | A=a] x E[X2 | A=a]. To do this you must:

1. Generate a set of candidate answers.
2. For each candidate answer, estimate E[X1 | A=a] and E[X2 | A=a] individually using your internal knowledge then compute O(a) for that candidate.
3. Return the answer a* that maximises O.

\end{spverbatim}

The template extends naturally to $n$ variables by adding further random variable definitions and adjusting $O(a)$ accordingly. An OR composition is obtained by replacing the product in $O(a)$ with a sum, and a NOT condition on a binary variable $X_i$ by replacing $\mathbb{E}[X_i \mid A=a]$ with $1 - \mathbb{E}[X_i \mid A=a]$. When variables exhibit a prerequisite dependency structure, each variable definition should additionally condition on the relevant parent variables being active, reflecting the binary gating structure of Section~\ref{ssec:binary-gating}.

\section{Experiments}\label{sec:experiments}

To validate our prompting framework, we run experiments on the MovieLens 1M dataset \cite{harper2015}. Given the first $100$ movies that a user has rated between $1$ and $5$, the LLM's task is to recommend the top $10$ movies for that user from the next $50$ in their watched list.

To cast the problem as a multi-objective task, we further condition that the user is only interested in movies in the comedy and romance genres. To ensure sufficient signal in the test set, we only select users that have at least $5$ movies in the test window that belong to both genres and carry a rating of $4$ or higher, and that have a total watch history of at least $150$ movies. We randomly select $20$ users and for each user we run $20$ queries to account for stochasticity in LLM outputs. We evaluate three prompt types:

\begin{enumerate}
\item \textbf{Basic}: The LLM is told that the user is in the mood for comedy and romance movies.
\item \textbf{Harsh}: The LLM is told that the user is only interested in comedy and romance movies and that it should not suggest anything outside of these genres.
\item \textbf{UtilityMax}: A prompt is constructed based on the template of Section~\ref{ssec:template} using multiplicative utility under conditional independence with three random variables - a categorical random variable $S$ for the predicted score, a binary random variable $G_1$ for the comedy genre, and a binary random variable $G_2$ for the romance genre. Therefore, the objective is:
\vspace{0.1cm}
\begin{equation}
O(a) = \mathbb{E}[S \mid A=a] \times P(G_1=1 \mid A=a) \times P(G_2=1 \mid A=a)
\end{equation}

\end{enumerate}

The LLM is not given the test movie scores, nor the genre labels for any movie in either the training or test set. Under UtilityMax, the LLM must therefore estimate $\mathbb{E}[S \mid A=a]$, $P(G_1=1 \mid A=a)$, and $P(G_2=1 \mid A=a)$ for each candidate movie using only its title and the user's training history.

The recommended list is evaluated against the test set, where a movie is considered a positive match only if it carries a rating of $4$ or higher and belongs to both the comedy and romance genres. We report Precision@10 and Normalised Discounted Cumulative Gain (NDCG@10) averaged across all users and runs. Experiments are conducted across three frontier models: Claude Sonnet 4.6, GPT-5.4, and Gemini 2.5 Pro. 

Our results are summarized in Table~\ref{tab:results}. UtilityMax outperforms both Basic and Harsh prompt types across all three 
models on both metrics. For example, with Claude Sonnet 4.6 we observe a $12.7\%$ and $16.5\%$ improvement in Precision@10 and NDCG@10 respectively over the Basic prompt, and a $11.9\%$ and $18.8\%$ improvement over the Harsh prompt.

\begin{table}[h]
\centering
\begin{tabular}{llcccc}
\toprule
\textbf{Model} & \textbf{Prompt} & \textbf{Precision@10} & \textbf{NDCG@10} \\
\midrule
\multirow{3}{*}{Claude Sonnet 4.6}
& Basic & 0.418 (0.054) & 0.570 (0.070) \\
& Harsh & 0.421 (0.067) & 0.559 (0.083) \\
& UtilityMax & \textbf{0.471 (0.060)} & \textbf{0.664 (0.064)} \\
\midrule
\multirow{3}{*}{GPT-5.4}
& Basic & 0.532 (0.052) & 0.739 (0.056) \\
& Harsh & 0.518 (0.050) & 0.712 (0.060) \\
& UtilityMax & \textbf{0.578 (0.047)} & \textbf{0.788 (0.050)} \\
\midrule
\multirow{3}{*}{Gemini 2.5 Pro}
& Basic & 0.417 (0.073) & 0.575 (0.098) \\
& Harsh & 0.449 (0.086) & 0.601 (0.108) \\
& UtilityMax & \textbf{0.496 (0.079)} & \textbf{0.667 (0.103)} \\
\bottomrule
\end{tabular}
\caption{Precision@10 and NDCG@10 averaged over $20$ users and $20$ queries per user. Standard deviations are shown in parentheses.}
\label{tab:results}
\end{table}

It is interesting to note that the Harsh prompt type does not consistently outperform Basic across models. Harsh underperforms
Basic on NDCG@10 for Claude Sonnet 4.6 (0.559 vs 0.570) and on both metrics for GPT-5.4. This inconsistency suggests that increasing the forcefulness of a natural language objective does not reliably resolve the underlying ambiguity in how multiple objectives are weighted. UtilityMax eliminates this ambiguity entirely through formal mathematical specification, which we believe accounts for its consistent superiority across all models and metrics.

GPT-5.4 achieves substantially higher absolute scores across all prompt types compared to the other two models. This may reflect training data overlap between GPT-5.4 and the MovieLens dataset, which is widely used and publicly available. However, the key finding for this paper is that UtilityMax continues to outperform both natural language baselines even for GPT-5.4, suggesting that the formal objective provides genuine additional signal regardless of model capability.

To assess statistical significance we apply a one-sided paired Wilcoxon signed-rank test on per-user mean NDCG@10, comparing UtilityMax against each baseline independently. The null hypothesis is that UtilityMax and the baseline produce equal NDCG scores; the alternative is that UtilityMax is superior. As shown in Table~\ref{tab:wilcoxon}, UtilityMax significantly outperforms both baselines across all three models ($p < 0.01$ in all cases).

\begin{table}[h]
\centering
\begin{tabular}{lcc}
\toprule
\textbf{Model} & \textbf{vs Basic} & \textbf{vs Harsh} \\
\midrule
Claude Sonnet 4.6 & 0.0028 & 0.0006 \\
GPT-5.4 & 0.0068 & 0.0042 \\
Gemini 2.5 Pro & 0.0014 & 0.0032 \\
\bottomrule
\end{tabular}
\caption{One-sided paired Wilcoxon signed-rank test p-values for UtilityMax vs each baseline on NDCG@10. All results are significant at the $p < 0.01$ level.}
\label{tab:wilcoxon}
\end{table}

Finally, we note that the effectiveness of UtilityMax depends on the underlying model's ability to produce well-calibrated probability estimates. A model that cannot reliably estimate the objective components may not benefit from this framework, and may even perform worse than natural language prompting. The results presented here suggest that current frontier models are capable enough to leverage the formal objective effectively, but we anticipate that weaker models may fall below this capability threshold. We leave a systematic investigation of this threshold to future work.

\section{Final Remarks}\label{sec:final-remarks}

This paper has introduced UtilityMax, a formal zero-shot prompting framework for multi-objective LLM tasks. The key insight is that by representing the task as a formal optimization objective, the LLM is constrained to reason explicitly about each component
of that objective, rather than relying on a natural language interpretation that may be ambiguous.

The effectiveness of UtilityMax depends on two key factors. First, the underlying model must be expressive enough to produce well-calibrated probability estimates for the objective components. This requires $K$ to be sufficiently rich -- drawing on the model's internal knowledge, externally provided context, or iterative dialogue. While UtilityMax requires no task exemplars and is therefore zero-shot in that sense, enriching $K$ through additional context or chat iteration can improve performance without altering the formal objective structure.

Second, the optimization objective must be carefully designed so that its variables capture the most important components of the underlying task. Including irrelevant variables introduces redundancy and increases computational cost, while omitting important variables will cause the framework to optimize a proxy that does not fully reflect the true goal. The designer must therefore exercise judgement in selecting variables that are both necessary and sufficient for the task at hand.

Several directions for future research present themselves:

\begin{enumerate}
\item Perhaps the most practically important is automating the construction of the UtilityMax prompt. Specifically, developing a method by which an LLM can extract the relevant variables from a natural language task description and format them into the UtilityMax framework automatically. This would remove the requirement for a human designer to manually specify the optimization objective.
\item A second direction is characterising further (DAG, utility) configurations admitting the decomposition $\mathbb{E}[U \mid A] = \sum_{k=1}^N \prod_{i=1}^n \mathbb{E}[f_{i,k}(X_i) \mid A]$. Conditional independence and binary gating are two sufficient conditions. Other sufficient conditions likely follow from results in probabilistic graphical models, but a unifying characterisation of the full admissible class remains open.
\item In the current framework, the LLM relies solely on $K$ to estimate each component probability from scratch. However, in many real-world settings historical data may be available that could be used to produce empirically calibrated prior estimates. Machine learning methods for parameter estimation in probabilistic graphical models could be applied to learn initial probability estimates for each node in the DAG. These data-driven priors would then serve as a starting point that the LLM refines using its broader knowledge and contextual understanding given its answer $A$. This is analogous to Bayesian updating where the data provides the prior and the LLM provides the likelihood update.
\item Finally, a systematic investigation of the capability threshold below which UtilityMax ceases to be beneficial would help practitioners identify which models are suitable for this framework.
\end{enumerate}

Overall, UtilityMax is a promising approach to improving LLM task performance through formal objective specification. Our experimental evidence across three frontier models demonstrates consistent improvements over natural language baselines, and we hope this work motivates further research into the formal specification of LLM objectives as a complement to existing prompt engineering techniques.

\bibliographystyle{unsrt}
\bibliography{references} 

\end{document}